\documentclass[runningheads]{llncs}
\usepackage{graphicx}
\usepackage{booktabs}
\begin{document}
\title{Transfer Learning of Transformer-based Speech Recognition Models from Czech to Slovak}
%
%

\author{
Jan Lehečka\inst{1}\orcidID{0000-0002-3889-8069} \and
Josef V. Psutka\inst{1}\orcidID{0000-0003-4761-1645} \and Josef Psutka\inst{1}\orcidID{0000-0002-0764-3207}
}

%

\institute{Department of Cybernetics, University of West Bohemia in Pilsen, Czech Republic\\
\email{\{jlehecka,psutka\_j,psutka\}@kky.zcu.cz}}

\maketitle              
\begin{abstract}
In this paper, we are comparing several methods of training the Slovak speech recognition models based on the Transformers architecture. Specifically, we are exploring the approach of transfer learning from the existing Czech pre-trained Wav2Vec 2.0 model into Slovak. We are demonstrating the benefits of the proposed approach on three Slovak datasets. 
Our Slovak models scored the best results when initializing the weights from the Czech model at the beginning of the pre-training phase.
Our results show that the knowledge stored in the Cezch pre-trained model can be successfully reused to solve tasks in Slovak while outperforming even much larger public multilingual models.

\end{abstract}
\section{Introduction}
Transfer learning in speech recognition has been shown to be effective in improving accuracy and reducing the amount of training data required for new tasks. It is especially useful in scenarios where the amount of available training data is limited, such as low-resource languages or domains with specific acoustic characteristics. The aim of this paper is to identify a suitable transfer learning approach for two languages, Czech and Slovak. These two languages have many similarities, both in their written form and pronunciation. 

In our experiments, we are comparing several methods of training the Slovak models for the target task of automatic speech recognition (ASR). Specifically, we are investigating the possibilities of transferring the knowledge from the existing pre-trained Czech model into Slovak ASR tasks. 
Since Czech and Slovak have a lot in common, we expect this transfer learning approach to be beneficial in the target Slovak tasks because it can reuse the already trained knowledge common to both languages while suppressing the non-Slovak information in favor of Slovak-specific knowledge during the transfer. In this paper, we investigate the benefits of this transfer learning approach.

We demonstrate the benefits of the proposed approach on three ASR datasets (described in detail in section \ref{sec:finetuning_data}). Two of the used datasets (CommonVoice and VoxPopuli) are public speech recognition datasets used very often for the benchmarking of ASR systems in many languages \cite{babu22_interspeech,radford2022whisper}. The third dataset, MALACH, is the Slovak portion of the very unique and challenging speech recognition dataset containing testimonies of eyewitnesses of the Holocaust recorded during 90'. We consider the MALACH dataset to be extremely important dataset for several reasons: (1) it preserves extremely valuable testimonies from our recent history, which should not be forgotten and which, alas, cannot be extended or scaled up anymore because the number of direct witnesses of the Holocaust rapidly decreases to zero as time goes on; (2) every improvement in the speech recognition accuracy unlocks new valuable historical and cartographical information encoded in the spoken utterances for researchers and public searching in this vast archive; (3) since most of the speakers were very old at the time of recording and the testimonies were spoken under heavy emotions, it is a challenging dataset to test the robustness, zero-shot performance and transfer learning ability of existing ASR models. 

\section{Transfer Learning from Czech to Slovak}

As mentioned above, Czech and Slovak share many similarities not only in their written form  but also phonetically. Czech orthography serves as a model for several other Balto-Slavic languages that use the Latin alphabet. Slovak can be regarded as its direct descendant from this perspective.   Both languages use comparable diacritics and have a similar, often interchangeable  relationship between letters and the sounds they represent. The significant similarity between the two languages can also be attributed to the fact that they were both official languages in the same country for over 40 years (in Czechoslovakia). In this article, we will focus only on the graphemic aspect of these languages. For a more detailed comparison of Czech and Slovak in the context of acoustic modeling, please refer to \cite{PsutkaJ_2005_Automatic_1,Nouza2010,10.1007/978-3-642-00525-1_24}.

In the Czech language, there are a total of 42 letters that are used. This includes the 26 letters of the basic Latin alphabet as well as 15 letters that have diacritical marks such as a caron [\v{}],  acute [\'{}], or a overring [\r{}]. In addition, there is a digraph [\textit{ch}] that represents a phoneme /x/ (SAMPA is used in all cases of phonetic notation \cite{SAMPA}) and is considered one of the letters of the Czech alphabet. There are two different ways to write a long /u:/ in Czech: [\textit{\'{u}}\,] and [\textit{\r{u}}\,], but they have the same pronunciation. One form cannot occur in the initial position, while the other occurs exclusively in the initial position or at the beginning of the root of a compound word.

The Slovak alphabet is the longest alphabet among Slavic and other European languages, consisting of a total of 46 letters. It includes the 26 letters of the basic Latin alphabet that are also used in Czech. Additionally, there are 17 letters that have diacritical marks, which include diaeresis [\"{}] and a circumflex [\^{}] but do not include a overring [\r{}]. But only five of these diacritical letters differ from those used in Czech ([\textit{\"{a}}\,] [\textit{\v{l}}\,] [\textit{\'{l}}\,] [\textit{\^{o}}\,] [\textit{\'{r}}\,]). Moreover, there are two additional digraphs present in the Slovak alphabet, i.e.  [\textit{dz}] and [\textit{d\v{z}}]. These letters represent phonemes /dz/ and /dZ/.

\section{Wav2Vec 2.0}
Wav2Vec 2.0 models have recently become a new state-of-the-art paradigm in ASR tasks outperforming the previous architectures by a large margin \cite{baevski2020wav2vec}.
It is a deep neural network pre-trained to reconstruct the corrupted audio signals. The model consists of a multi-layer convolutional neural network (referred to as a feature encoder) followed by a multi-layer Transformer encoder \cite{vaswani2017attention}.
The convolutional feature encoder processes the raw input signal and produces a sequence of latent-speech representations. Each of these latent-speech representations is a vector encoding one 20ms-long frame of the input signal with only a small (5ms) context being taken into account. The attention-based Transformer then converts latent-speech representations into contextualized speech representations while paying attention to the full context of the input signal.

The training of Wav2Vec models consists of two phases: self-supervised pre-training and supervised fine-tuning. 
The phase of self-supervised pre-training requires a large-scale unlabeled speech dataset, from which the model learns the contextualized speech representations by predicting masked frames.
Moreover, the model is pre-trained also to solve a contrastive task over quantized speech representations, so the model is forced to map input frames into discrete speech units and correctly identify masked frames among a set of distractors.
During this phase, the model does not have any orthographical information about the processed speech as it has access only to the raw audio signal, so it is pre-trained to catch and encode the meaning of individual audio frames only based on its context.

The pre-training phase is essential to equip the model with deep knowledge mined from tens of thousands of hours of unlabeled speech. This knowledge constitutes a great advantage over models trained from scratch using labeled data only. From this point of view, the pre-trained weights of the Wav2Vec model could be seen as a very clever initialization of the model weights for supervised training. In this paper, we are investigating the benefits of clever initialization also for the pre-training, i.e., not starting from random weights from scratch but using weights of a model pre-trained from much more speech data from a language that is somehow similar. This way, the model could preserve the information common to both languages and reuse it when solving tasks in the other language. 

After the pre-training is done, the model transfers the pre-trained knowledge into the target ASR task within the fine-tuning phase. This is a supervised phase requiring the training speech dataset to be labeled. In order to decode the most probable sequences of graphemes, the model is additionally equipped with a final Connectionist Temporal Classification (CTC) layer \cite{graves2006connectionist}. 
CTC is an alignment-free method for grouping audio frames belonging to the same output token in order to convert a sequence of frame-level predictions into a much shorter sequence of output tokens.
The CTC classification process can be described -- in a simplified way -- in 3 steps: 

\begin{enumerate}
    \item Assign the most probable output token to each audio frame.
    \item Group sub-sequences with the same token into a single token.
    \item Remove blank tokens.
\end{enumerate}

Tokens could be any speech or language units, e.g., phonemes, graphemes, sub-word units, words, etc. In this paper, we experimented with grapheme-based predictions, i.e., we predicted the sequence of characters. We chose the grapheme-based output units because it has several advantages: (1) the fine-tuned model works with very small vocabulary (the size of the alphabet plus several special tokens), so the decoding is fast, (2) it avoids out-of-vocabulary problems (any sequence of graphemes can be predicted), and (3) it can be used as a stand-alone full-fledged end-to-end speech recognizer without any additional postprocessing. 

\section{Experimental Setup}
In our experiments, we used existing pre-trained Wav2vec models or -- when not available -- we pre-trained new ones. We fine-tuned all pre-trained models on train and development parts of three Slovak ASR datasets. After that, we evaluated all models on the test part of relevant datasets. The test parts were held out during the whole fine-tuning process and had no speaker overlaps with train or development parts. 
We used implementation from \texttt{Fairseq} tool \cite{ott2019fairseq} for both pre-training and fine-tuning of models.

\subsection{Pre-trained Models}
In this section, we present all the pre-trained models we were experimenting with. We used three monolingual pre-trained Wav2Vec 2.0 models of the base size: Czech (denoted as \texttt{W2V2-cs}), Slovak (\texttt{W2V2-sk}), and a model transferred from Czech to Slovak (\texttt{W2V2-cs-sk}). To test the monolingual models against multilingual models, we also evaluated two popular large-scale multilingual models (Wav2Vec XLS-R and Whisper).
We are listing the models along with detailed information in the rest of this section.

\subsubsection{W2V2-cs}
The \texttt{W2V2-cs} is a monolingual model pre-trained solely from the Czech speech. We used the publicly available model \texttt{ClTRUS}\footnote{\url{https://huggingface.co/fav-kky/wav2vec2-base-cs-80k-ClTRUS}} \cite{wav2vec2-base-cs-80k-ClTRUS}. It has been trained from 80 thousand hours of Czech speech from various domains, mainly from the VoxPopuli dataset \cite{wang-etal-2021-voxpopuli} and records from Czech TV and radio shows.

\subsubsection{W2V2-sk}
The \texttt{W2V2-sk} is a monolingual model pre-trained solely from the Slovak speech. We didn't find any suitable public model, so we pre-trained a new base-sized model from scratch.
Since Transformer-based models are known to scale well with the size of pre-training data, we tried to gather as much public unlabeled speech data as possible. We collected over 17 thousand hours of Slovak speech from various sources.
The collection includes recordings from the Slovak portion of the VoxPopuli dataset \cite{wang-etal-2021-voxpopuli} (12k hours),
a mix of self-crawled records from Slovak TV shows (4.5k hours), 
the MALACH dataset (800 hours) and the Slovak portion of CommonVoice corpus 13.0 \cite{commonvoice:2020} (24 hours).
We used Wav2Vec 2.0 architecture \cite{baevski2020wav2vec} and adopted the same hyperparameter setting as in the paper, i.e., we trained the base model (12 Transformer blocks, model dimension 768, 8 attention heads, and a total of 95 million parameters) for 400 thousand steps with a batch size of about 1.6 hours. 
The pre-training took four days on a machine with eight NVIDIA A100 GPUs.

\subsubsection{W2V2-cs-sk}
The \texttt{W2V2-cs-sk} is a monolingual Slovak model which was not initialized randomly from scratch but rather from weights of the Czech model \texttt{W2V2-cs}. 
After the initialization, we pre-trained the model with the exact same setting and data as \texttt{W2V2-sk}. Thus, the only difference between \texttt{W2V2-sk} and \texttt{W2V2-cs-sk} is the initialization of weights.
We expect this model to identify, preserve and transfer the useful knowledge common to both languages while suppressing the non-Slovak information in favor of Slovak-specific knowledge during the pre-training. In this paper, we are exploring if and how much this transfer learning approach is beneficial. We are releasing this pre-trained Slovak model publicly to the research community\footnote{\url{https://huggingface.co/fav-kky/wav2vec2-base-sk-17k}}.

\subsubsection{W2V2-XLS-R-300M}
To compare monolingual models also with popular multilingual public models, we selected Wav2Vec XLS-R \cite{babu22_interspeech} as a  representative of large-scale pre-trained cross-lingual models. The model was pre-trained on approximately 436 thousand hours of unlabeled speech data from 128 languages (including both Czech and Slovak). We experimented with the 300M variant, which has more than 300 million parameters, i.e., more than $3\times$ more than the base Wav2Vec 2.0 model. We denote this model \texttt{W2V2-XLS-R-300M}.

\subsubsection{Whisper-large}
Finally, we compared our models with \texttt{Whisper-large} \cite{radford2022whisper}, another popular model trained on 99 languages (including both Czech and Slovak) from 680,000 hours of multilingual and multitask labeled data. This model differs from Wav2Vec models in two main aspects: (1) it is not an encoder-only model but has also a decoder serving as an audio-conditioned built-in language model, (2) the input is Mel spectrogram instead of the raw audio signal. We experimented with the large size of the model with 32+32 Transformer layers, dimension 1280, 20 attention heads, and a total of 1.55 billion trainable parameters. When decoding, we specified the language to Slovak, so the model didn't have to identify the language automatically from the input signal. As this model has already been fine-tuned on a large palette of datasets and tasks by authors, we didn't further fine-tune the model, and we used the downloaded weights directly.

\subsection{Fine-tuning}
\label{sec:finetuning}
We prepared all training and development ASR data consistently for all datasets. Where necessary, we sliced long training audio signals on speech pauses not to exceed the length of 30\,s. Longer utterances were discarded due to the memory limits of used GPUs during fine-tuning.
We removed non-speech events and punctuation from the transcripts and mapped all words into lowercase. 
We fine-tuned all models with the same setting as the base model in \cite{baevski2020wav2vec}, i.e., we trained for 80 thousand steps with a batch size of about 26 minutes per step, and the learning rate warmed up over the first 8\,000~steps to a maximum value of $2\times10^{-5}$, where it was held for the next 32\,000~steps, and finally decayed exponentially to zero. The weights of the feature encoder were frozen for the first 10\,000~steps of the fine-tuning.

\subsection{Fine-tuning Datasets}
\label{sec:finetuning_data}
We experimented with three datasets described in detail in the rest of this section. The statistics about individual datasets are tabulated in Tab.~\ref{tab:ASRstats}.

\setlength{\tabcolsep}{0.3em}
{ 
\begin{table}[htb]
  \caption{Fine-tuning datasets. We show the total number of speech hours, the number of utterances, and the total number of words in transcripts (in thousands).}
  \label{tab:ASRstats}
  \centering
  \begin{tabular}{lrrrcrrrcrrr}
    \toprule
     & \multicolumn{3}{c}{CommonVoice} & & \multicolumn{3}{c}{VoxPopuli} & & \multicolumn{3}{c}{MALACH} \\
    \cline{2-4} \cline{6-8} \cline{10-12} 
     & train & dev & test & & train & dev & test & & train & dev & test \\
    \midrule
    
    \textit{\# hours of audio} &   14.2 & 2.9 & 3.1 &  & 29.2 & 1.9 & 1.7 & & 94.3 & 2.0 & 1.2 \\
    \textit{\# utterances} &      13\,122 & 2\,474 & 2\,552  &  & 10\,410 & 664 & 604 & & 13\,160 & 273 & 500 \\
    \textit{\# words (in thousands)} &      48.0 & 11.0 & 10.2  &  & 233.2 & 14.6 & 13.4 & & 645.8 & 14.0 & 8.3 \\
   
    \bottomrule
  \end{tabular}
\end{table}
}

\subsubsection{CommonVoice} 
The CommonVoice dataset is a Slovak portion of the crowdsourced project Mozilla Common Voice \cite{commonvoice:2020}. We used corpus version 13.0, containing 20 hours of validated speech. We decided to keep also sentences reported as \textit{difficult pronunciation} in our training data. All other reported sentences (e.g., \textit{grammar or spelling}, \textit{different language} etc.) were ignored.

\subsubsection{VoxPopuli}
The VoxPopuli dataset \cite{wang-etal-2021-voxpopuli} is a large-scale multilingual speech corpus collected from 2009-2020 European Parliament event recordings. The Slovak portion contains $12.1$ thousand unlabeled hours and 35 hours with transcription. We ignored all train and development utterances without the raw transcription, decreasing the amount of transcribed data to $32.8$ hours.

\subsubsection{MALACH}
\label{sef:MALACH}
The Malach Archive preserves the memories of Holocaust survivors through audiovisual interviews in 32 languages. The recordings are characterized by natural speech with emotional outpourings and heavy accents due to the advanced age of the speakers (around 75 years old). Transfer learning can significantly increase recognition accuracy for such type of data, as it is difficult to find additional suitable data for acoustic modeling due to the nature of the corpus (more details can be found in \cite{malach}).

The Czech portion of the Malach data was released by the LDC in 2014 \cite{LDC2014S04}, comprising 400 randomly selected testimonies for training acoustic models. However, due to the manual transcription of only 15-minute segments of each testimony, the acoustic modeling process had access to only 100 hours of Czech speech data. Theoretically, the available data could contain up to 800 speakers. The Slovak section of the Malach corpus was transcribed similarly to the Czech section, with 15-minute segments of 400 testimonies transcribed for training. Additionally, 20 testimonies (10 men and 10 women) were fully transcribed to create the development and test portions of the Slovak corpus. In order to maintain consistency with other corpora and ensure a manageable test size, the size of the test set was limited to a reasonable level. A carefully selected subset of the transcribed data consisting of 500 sentences was utilized. To enhance the reliability of the results, all segments containing crosstalks were deliberately excluded from the test set, as they could potentially impact the findings. Therefore, this subset consisted only of continuous segments where either the survivor or the interviewer spoke, with no interruption or overlap from the other speakers.

\subsection{Decoding}
When transcribing the speech from fine-tuned models, we experimented with two decoding strategies: (1) using only the fine-tuned Wav2Vec model as a stand-alone end-to-end speech recognizer and (2) CTC beam search decoder using additional language information from a language model (LM) during the decoding. 
The decoding with strategy (2) usually improves speech recognition performance by bringing useful language information into the decoding process while penalizing improbable outputs in the target language. 

For strategy (2), we trained one large-scale general-purpose n-gram LM to be used in all experiments for all datasets. As training data, we used web pages from the Common Crawl project\footnote{\url{https://commoncrawl.org}}. We downloaded and processed 34 crawls from August 2018 to October 2021 following the same cleaning and deduplicating rules as in the English C4 dataset \cite{2019t5}. Together, we collected about 37GB of cleaned and deduplicated Slovak text containing 5.6 billion words from more than 16 million web pages. To keep the LM of a practical size, we pruned all unigrams with counts lower than ten and higher-order n-grams with counts lower than 100. We trained the LM in lowercase as all fine-tuning transcripts were converted into lowercase. The final LM contained 2.5 million unigrams and 12 million n-grams in total. We used \texttt{KenLM} \cite{heafield2011kenlm} toolkit to train the LM and \texttt{pyctcdecode}\footnote{\url{https://github.com/kensho-technologies/pyctcdecode}} tool to decode transcripts.  

\subsection{Evaluation}
We compared models in terms of word error rate (WER). Since all transcripts were cleaned from punctuation and cast into lowercase before the fine-tuning, our fine-tuned models cannot predict punctuation or upper-cased characters, so we did not consider casing and punctuation differences with the reference as errors.

Note that although our models are not able to predict cased transcriptions nor punctuation, which usually makes the transcript difficult to read, we are, in all relevant applications, applying also a postprocessing phase on generated transcripts, in which a specially trained transformer-based large language model restores the casing and punctuations in the transcripts. We found this approach more beneficial than training the Wav2Vec models to predict directly cased words and punctuation for two reasons: (1) the text-based language model is more accurate in this task as it can work with larger context and have a better understanding of the syntax and semantics of the spoken words, and (2) the training of Wav2Vec models is less confusing because both cased words and punctuation tokens do not correspond to any distinguishable acoustic units and yet, they would have different target labels.

\section{Results}
The results of our experiments are tabulated in Tab.~\ref{tab:results} (results with stand-alone Wav2Vec models) and  Tab.~\ref{tab:resultsLM} (results with Wav2Vec models using the language model in the decoder). When comparing corresponding values from both tables, we can confirm that including LM from Common Crawl into the CTC decoder significantly improves the ASR results for all models across all datasets.

\setlength{\tabcolsep}{0.6em}
\begin{table}[htbp]
\centering
\caption{Evaluation results in terms of WER $[\%]$ scored by end-to-end grapheme-based models. These results show how individual fine-tuned Transformer models perform when used as a stand-alone ASR system without any language model involved.}
\label{tab:results}
\begin{tabular}{lcrrr}
\toprule
 & \#params & \multicolumn{3}{c}{fine-tuned and evaluated on} \\
 \cline{3-5}
 & [in millions] & CommonVoice & VoxPopuli & MALACH \\ 
\midrule
\texttt{W2V2-cs} & 95 & 13.85 & 11.58 & 14.81 \\
\texttt{W2V2-sk} & 95 & \textbf{10.62} & 10.09 & 13.60 \\
\texttt{W2V2-cs-sk} & 95 & 10.95 & \textbf{9.76} & \textbf{13.30} \\
\midrule
\texttt{W2V2-XLS-R-300M} & 300 & \textbf{9.44} & 10.39 & 15.12 \\
\bottomrule
\end{tabular}
\end{table}

In the first row of both tables, we show the results of the Czech model \texttt{W2V2-cs} fine-tuned on the Slovak datasets. When compared with results in the second row from the Slovak model \texttt{W2V2-sk}, we can clearly see the Slovak model is better (which is expected), but moreover, we see that the difference is, in many cases, not so large (from 0.5\% to 3.2\% in terms of absolute WER reduction). This closeness confirms that Czech and Slovak have a lot in common, and we could get a reasonably good Slovak ASR system just by fine-tuning the Czech pre-trained model on a small amount of Slovak labeled speech. The larger the fine-tuning dataset is, the smaller the difference between the performance of the Czech and Slovak pre-trained models is.

Now, let's concentrate on the differences between the second row (Slovak model \texttt{W2V2-sk} pre-trained from scratch from the Slovak-only speech) and the third row (Slovak model \texttt{W2V2-cs-sk} initialized from the Czech model before pre-training). For two datasets (VoxPopuli and MALACH), we can observe a small but consistent decrease in WER gained by this transfer learning. However, for the CommonVoice dataset, we got the best results (among the base-sized models) from the pure Slovak model. After an analysis of the errors, we believe this is caused by an insufficient amount of training data. There are just 14.2 hours of labeled Slovak speech in the training CommonVoice dataset. We observed many Czech forms of Slovak words in the transcripts from the \texttt{W2V2-cs-sk} model fine-tuned on the CommonVoice dataset, indicating that the model still has a lot of the original Czech-related knowledge even after the transfer to Slovak and that this amount of train labeled data is not enough to override the Czech-related knowledge in the model.

\setlength{\tabcolsep}{0.6em}
\begin{table}[htbp]
\centering
\caption{Evaluation results in terms of WER $[\%]$ scored by models also incorporating the language model in the decoder. These results show how individual fine-tuned Transformer models perform when also adding the language model probabilities into the decoding process. Values decorated with an asterisk (*) are scored by a general-purpose ASR model without fine-tuning to the target dataset.}
\label{tab:resultsLM}
\begin{tabular}{lcrrr}
\toprule
 & \#params & \multicolumn{3}{c}{fine-tuned and evaluated on} \\
 \cline{3-5}
 & [in millions] & CommonVoice & VoxPopuli & MALACH \\ 
\midrule
\texttt{W2V2-cs} & 107 & 11.25 & 10.04 & 12.79 \\
\texttt{W2V2-sk} & 107 & \textbf{8.68} & 9.02 & 12.32 \\
\texttt{W2V2-cs-sk} & 107 & 8.82 & \textbf{8.88} & \textbf{11.57} \\
\midrule
\texttt{W2V2-XLS-R-300M} & 312 & \textbf{6.90} & 9.09 & 12.17 \\
\midrule
\texttt{Whisper-large} & 1\,550 & *34.61 & *19.30 & *27.49 \\ 
\bottomrule
\end{tabular}
\end{table}

The multilingual \texttt{W2V2-XLS-R-300M} scored the best result among all models on the CommonVoice dataset. We attribute this result to the fact that it was pre-trained on the whole CommonVoice dataset containing 7 thousand hours containing similar sentences (the domain of CommonVoice is a read speech primarily from Wikipedia sentences) in various languages. Thus, the pre-trained embeddings could better encode information in this dataset than other models, where the CommonVoice dataset was only a very small part of the pre-training corpus. 
However, although more than $3\times$ larger, it did not perform better on the other two datasets, for which our smaller monolingual models performed slightly (VoxPopuli dataset) or significantly (MALACH dataset) better. 

Finally, the results from the Whisper model are far from all fine-tuned models. Although this model was not directly fine-tuned on the target datasets, CommonVoice and VoxPopuli datasets were a part of the huge labeled training dataset of the model. These results, which correspond to the reported results in \cite{radford2022whisper}, suggested that general-purpose models -- even the huge ones -- do not always perform well on low-resources languages and tasks.

To sum up our results, the transfer learning between Czech and Slovak is, in most cases, beneficial, and the more labeled data for the target domain there is, the more we can benefit from this transfer by reusing the knowledge common to both languages. We also showed that monolingual models pre-trained on a single language can successfully compete with the much larger multilingual models.

\section{Conclusion}
In this paper, we compared several methods of training the Slovak ASR models and evaluated the models on three Slovak datasets. Our results showed that the proposed transfer learning approach from the Czech pre-trained model can bring significant reduction in terms of speech recognition WER, especially when the fine-tuning dataset is large enough.

Our base Wav2Vec 2.0 models performed better on two datasets (including the extremely important MALACH dataset) than $3\times$ larger Facebook's XLS-R model and much better on all three datasets than $16\times$ larger OpenAI's Whisper model.
Since such a reduction of the model size while preserving or improving the performance could save a lot of energy required for the inference, we release the pre-trained Slovak model publicly for the research community.

\subsubsection*{Acknowledgments.}
This research was supported by the Ministry of the Interior of the Czech Republic, project No. VJ01010108. 
Computational resources were provided by the e-INFRA CZ project (ID:90254), supported by the Ministry of Education, Youth and Sports of the Czech Republic.

\bibliographystyle{splncs04}
\bibliography{main}

\end{document}